\documentclass[conference]{IEEEtran}
\IEEEoverridecommandlockouts
\usepackage{cite}
\usepackage{amsmath,amssymb,amsfonts}
\usepackage{algorithmic}
\usepackage{graphicx}
\usepackage{subcaption} 
\usepackage{textcomp}
\usepackage{xcolor}
\usepackage{booktabs}
\usepackage{url}
\usepackage{listings}
\def\BibTeX{{\rm B\kern-.05em{\sc i\kern-.025em b}\kern-.08em
    T\kern-.1667em\lower.7ex\hbox{E}\kern-.125emX}}

\begin{document}

\title{COA-GPT: Generative Pre-trained Transformers for Accelerated Course of Action Development in Military Operations
\thanks{This research was sponsored by the Army Research Laboratory and was accomplished under Cooperative Agreement Number W911NF-23-2-0072. The views and conclusions contained in this document are those of the authors and should not be interpreted as representing the official policies, either expressed or implied, of the Army Research Laboratory or the U.S. Government. The U.S. Government is authorized to reproduce and distribute reprints for Government purposes notwithstanding any copyright notation herein.\\\\
This paper was originally presented at the NATO Science and Technology Organization Symposium (ICMCIS) organized by the Information Systems Technology (IST) Panel, IST-205-RSY - the ICMCIS, held in Koblenz, Germany, 23-24 April 2024.}
}

\author{\IEEEauthorblockN{Vinicius G. Goecks}
\IEEEauthorblockA{\textit{DEVCOM Army Research Laboratory}\\
Aberdeen Proving Ground, Maryland, USA\\
vinicius.goecks@gmail.com}
\and
\IEEEauthorblockN{Nicholas Waytowich}
\IEEEauthorblockA{\textit{DEVCOM Army Research Laboratory}\\
Aberdeen Proving Ground, Maryland, USA\\
nicholas.r.waytowich.civ@army.mil}
}

\maketitle

\begin{abstract}
The development of Courses of Action (COAs) in military operations is traditionally a time-consuming and intricate process. Addressing this challenge, this study introduces COA-GPT, a novel algorithm employing Large Language Models (LLMs) for rapid and efficient generation of valid COAs. COA-GPT incorporates military doctrine excerpts and domain expertise to LLMs through in-context learning, allowing commanders to input mission information – in both text and image formats – and receive strategically aligned COAs for review and approval. Uniquely, COA-GPT not only accelerates COA development, producing initial COAs within seconds, but also facilitates real-time refinement based on commander feedback. This work evaluates COA-GPT in a military-relevant scenario within a militarized version of the StarCraft II game, comparing its performance against an expert human and state-of-the-art reinforcement learning algorithms. Our results demonstrate COA-GPT's superiority in generating strategically sound COAs more swiftly, with the added benefits of enhanced adaptability and alignment with commander intentions. COA-GPT's capability to rapidly adapt and update COAs during missions presents a transformative potential for military planning, particularly in addressing planning discrepancies and capitalizing on emergent windows of opportunity.
Performance videos of our method can be seen at \url{https://sites.google.com/view/coa-gpt}.

\end{abstract}


\section{Introduction}

\begin{figure*}[htbp]
    \centerline{\includegraphics[width=0.95\textwidth]{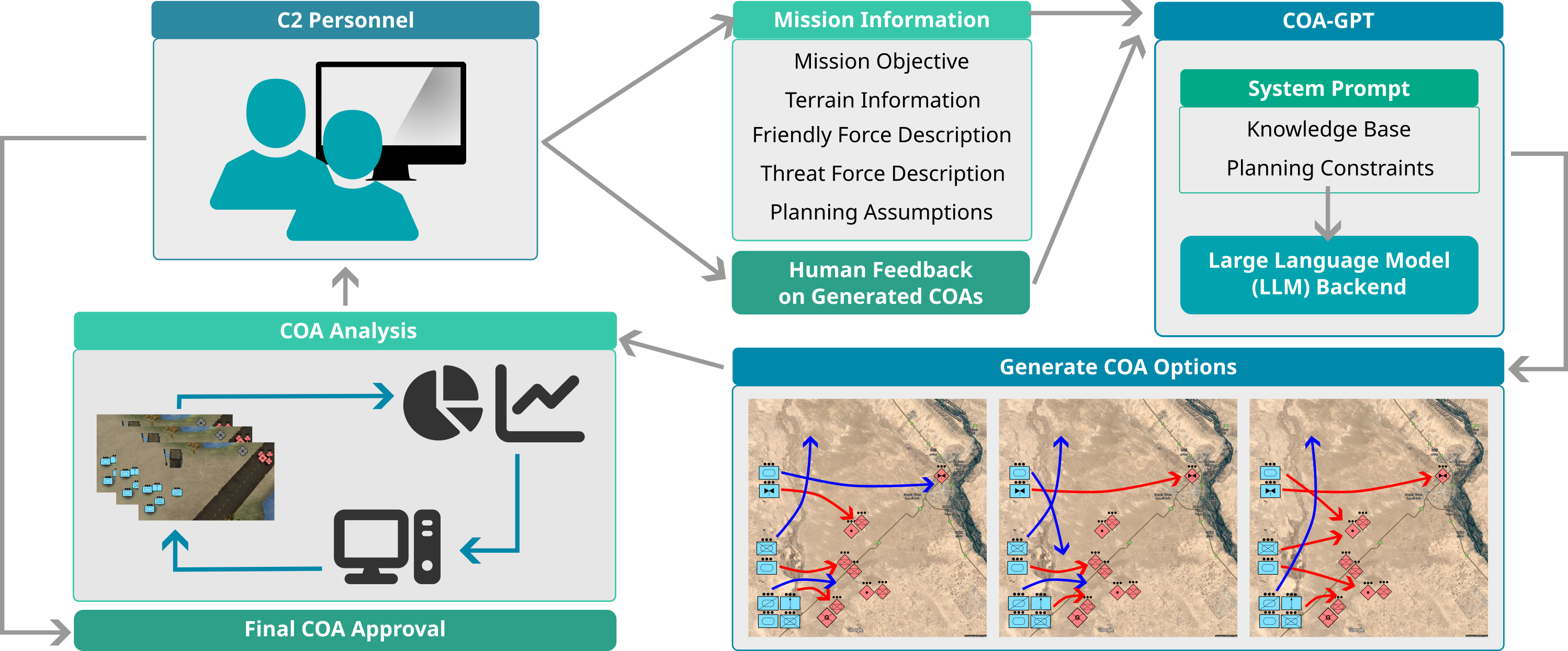}}
    \caption{Overview of the proposed method for COA-GPT. COA-GPT consists of a large language model (LLM) that learns via in-context learning by initially being prompted with a knowledge base and additional constraints to be respected during COA development. Command and control personnel supply mission information, which is used by COA-GPT to generate options for COAs and iterates with humans via natural language until the final COA is selected.}
    \label{fig:coa-gpt_diagram}
\end{figure*}

The future battlefield presents an array of complex and dynamic challenges for Command and Control (C2) personnel, requiring rapid and informed decision-making across multifaceted domains. As warfare continues to evolve, integrating and synchronizing diverse assets and effects across air, land, maritime, information, cyber, and space domains becomes increasingly critical.
The C2 operations process, encompassing planning, preparation, execution, and continuous assessment, must adapt to these complexities while dealing with real-time data integration and operating under conditions of Denied, Degraded, Intermittent, and Limited (DDIL) communication \cite{army2019ADP5-0, army2022FM6-0}.

In this high-stakes environment, maintaining decision advantage – the ability to make timely and effective decisions faster than adversaries – is paramount. The Military Decision Making Process (MDMP)~\cite{army2022FM5-0}, a cornerstone of C2 planning~\cite{marr2001military}, faces the pressing need to evolve with the accelerating pace and complexity of modern warfare \cite{farmer2022four}.
This necessitates executing the MDMP within increasingly shorter timescales to exploit fleeting opportunities and respond adaptively to the dynamic conditions of the battlefield.

The development of Courses of Action (COAs), a core element of decision-making, highlights these challenges. Traditionally, COA development is meticulous and time-intensive, relying heavily on military expertise. The demands of modern warfare require more efficient COA development and analysis methods.

Enter Large Language Models (LLMs), a transformative technology in the field of natural language processing \cite{devlin2018bert, raffel2020exploring, brown2020language}. LLMs have shown immense potential in various applications, including disaster response \cite{ningsih2021disaster, zhou2022victimfinder, ghosh2022gnom} and robotics \cite{ahn2022can, huang2022inner, mees2023grounding}, by processing extensive data to generate human-like text from given prompts. This research paper proposes COA-GPT, a framework that explores the application of LLMs to expedite COA development in military operations via in-context learning.

COA-GPT leverages LLMs to swiftly develop valid COAs, integrating military doctrine excerpts and domain expertise directly into the system's initial prompts. Commanders can input mission specifics and force descriptions – in both text and image formats - receiving multiple, strategically aligned COAs in a matter of seconds.

Our main contributions are:
\begin{itemize}
    \item COA-GPT, a novel framework leveraging LLMs for accelerated development and analysis of COAs, integrating military doctrine excerpts and domain expertise effectively.
    \item Empirical evidence showing that COA-GPT outperforms existing baselines in generating COAs, both in terms of speed and alignment with military strategic goals.
    \item Demonstration of COA-GPT's ability to improve COA development through human interaction, ensuring that generated COAs align closely with commanders' intentions and adapt dynamically to battlefield scenarios.
\end{itemize}

\section{Military Relevance}

The current C2 processes in military operations are predominantly linear and involve numerous sequential steps, which can be cumbersome and slow in the fast-paced environment of modern warfare \cite{army2022FM5-0, army2022FM6-0}. The advent of AI in military planning, particularly the COA-GPT system, presents an opportunity to radically streamline these processes.

COA-GPT significantly enhances the MDMP by enabling the concurrent development and analysis of COAs. This facilitates quicker decision-making and enhances operation control, leading to the swift generation of actionable intelligence for informed command decisions.
Integrating AI tools like COA-GPT in C2 processes can reduce military operations' physical footprint, lessening the need for extensive personnel and logistics, and aligning with the vision of distributed command structures.

Solutions such as COA-GPT can integrate with war gaming simulators and battlefield sensor data in real-time, enabling quick adaptation to battlefield dynamics. This approach emphasizes early COA analysis integration for swift optimization and comparison~\cite{farmer2022four}.
Additionally, by integrating human feedback, COA-GPT allows personnel to adapt AI-generated COAs with their expertise and situational awareness for informed final selections.
C2 personnel dynamically interact with COA-GPT to specify objectives, input data, set planning constraints, and adjust proposed COAs, ensuring decisions align with strategic intent and situational requirements.

\section{Related Work}

\subsection{Planning with Large Language Models}

The integration of Large Language Models (LLMs) in the plan of action development has been applied to various sectors, including disaster response operations.
Similar to military operations, disaster response demands rapid, informed decision-making under severe time constraints and high-pressure conditions \cite{Rennemo2014ATS, Jayawardene2021TheRO, Uhr2018AnES}. Traditionally, action plan development in such contexts has been a laborious process, heavily reliant on the experience and expertise of the personnel involved. Given the stakes involved, where delays can result in loss of life, there is a critical need for more efficient and reliable plan development methodologies \cite{kovel2000modeling, alsubaie2013platform, Rennemo2014ATS, SUN2021107213}.
In the realm of disaster response, the DisasterResponseGPT~\cite{goecks2023disasterresponsegpt} algorithm leverages the capabilities of LLMs to quickly generate viable action plans, integrating essential disaster response guidelines within the initial prompt.



The recent studies by Ahn et al. (2022)~\cite{ahn2022can} and Mees et al. (2023)~\cite{mees2023grounding} expand the scope of LLM applications to include robotic and visual affordances, illustrating how LLMs can be grounded in real-world contexts.
Continuing to show the potential of LLMs for planning, the introduction of Voyager by Wang et al. (2023)~\cite{wang2023voyager}, an LLM-powered embodied lifelong learning agent in Minecraft, represents an important step in autonomous, continual learning and skill acquisition in complex, ever-changing environments. Similarly, the STEVE-1~\cite{lifshitz2023steve} model showcases the potential of LLMs in guiding agents to follow complex instructions in virtual settings, leveraging advancements in text-conditioned image generation and decision-making algorithms.

The integration of LLMs into planning has significantly advanced rapid decision-making capabilities in a diverse range of scenarios.
COA-GPT builds upon these foundations and demonstrates the versatility of LLMs in translating strategic concepts into actionable plans for military operations.

\subsection{AI for Military Planning and Operations}

The application of AI and LLMs in military planning and operations is a field of growing interest and significant potential. The First-Year Report of ARL Director's Strategic Initiative, focusing on Artificial Intelligence for Command and Control (C2) of Multi-Domain Operations~\cite{dsifirstyear}, exemplifies this trend.
The report discusses ongoing research into whether AI could support agile and adaptive C2 in multi-domain forces.

Exploring the synergy between gaming platforms and military training, Goecks et al. (2023) provide insight into how AI algorithms, when combined with gaming and simulation technologies, can be adapted to replicate aspects of military missions~\cite{goecks2023games}.
In a similar vein, Waytowich et al. (2022)~ \cite{waytowich2022learning} demonstrate the application of deep reinforcement learning (DRL) in commanding multiple heterogeneous actors in a simulated command and control task, modeled on a military scenario within StarCraft II. Their findings indicate that agents trained via an automatically generated curriculum can match or even surpass the performance of human experts and state-of-the-art DRL baselines.

In addition to these developments, Schwartz (2020)~\cite{schwartz2020ai} delves into the application of AI in the Army's MDMP, specifically in the COA Analysis phase.
They demonstrate how AI can assist commanders and their staff in quickly developing and optimizing multiple courses of action in response to the complexities of modern, hyper-contested battlefields.
The study also highlights the increasing importance of Multi-Domain Operations (MDO) and the challenges presented by near-peer adversaries equipped with advanced Anti-Access Area Denial (A2AD) capabilities. 

In contrast to these AI-driven approaches, traditional military planning processes, as outlined in U.S. Army's ATP 5-0.2~\cite{army2022ATP5-0.2-1}, involve comprehensive guidelines for staff members in large-scale combat operations.
While providing a consolidated source of key planning tools and techniques, these traditional methodologies often lack the agility and adaptability offered by modern AI systems in rapidly changing combat environments.

In the evolving landscape of military planning and operations, the integration of AI and LLMs signifies a shift towards more agile and adaptive command and control strategies.
COA-GPT exemplifies this shift, leveraging LLMs for accelerated COA development, thus addressing the limitations of manual planning processes and previous automated approaches.

\section{Methods and Experiments}\label{sec:methods}

In this research, we leveraged the in-context learning capabilities of LLMs to create COA-GPT, a virtual assistant designed to efficiently generate COAs for military operations.

COA-GPT is prompted to understand that it serves as a military commander's assistant to aid C2 personnel in developing COAs.
It is aware that its inputs will include mission objectives, terrain information, and details on friendly and threat forces as provided by the commander in text and/or image format.
It is also instructed to use specific commands to assign tasks for each asset in the friendly forces.
Furthermore, COA-GPT has access to military doctrine excerpts, covering forms of maneuver (envelopment, flank attack, frontal attack, infiltration, penetration, and turning movement), offensive tasks (movement to contact, attack, exploitation, and pursuit), and defensive tasks (area defense, mobile defense, and retrograde).
The complete system prompt and doctrinal excerpts given to COA-GPT are reproduced in Appendix~\ref{appendix:system_prompt}.

As depicted in Figure \ref{fig:coa-gpt_diagram}, the COA-GPT assistant communicates with C2 personnel via natural language. It receives mission-related information such as objectives, terrain details, friendly and threat force description and arrangement, and any planning assumptions the C2 staff might have.
For the LLM in the back end, we use OpenAI's GPT-4-Turbo (named ``\textit{gpt-4-1106-preview}" in their API) for text-only experiments and GPT-4-Vision (named ``\textit{gpt-4-vision-preview}") for tasks where mission information is given in both text and image format.

Upon receiving this information, COA-GPT generates several COA options, each with a designated name, purpose, and visual representation of the proposed actions. Users can select their preferred COA and refine it through textual suggestions. COA-GPT processes this feedback to fine-tune the selected COA. Once the commander approves the final COA, COA-GPT conducts an analysis and provides performance metrics. 

The generation of COAs by COA-GPT is remarkably swift, completing in seconds. Including the time for commander interaction, a final COA can be produced in just a few minutes. This efficiency underscores COA-GPT's potential to transform COA development in military operations, facilitating rapid adjustments in response to planning phase discrepancies or emerging opportunities.


\subsection{Scenario and Experimental Setup}

\begin{figure}[htbp]
    \centerline{\includegraphics[width=0.8\columnwidth]{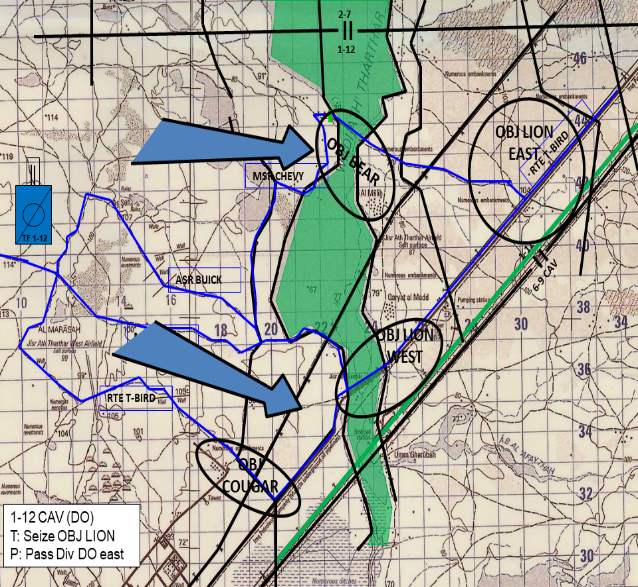}}
    \caption{TF 1‒12 CAV Area of Operations in TigerClaw~\cite{dsifirstyear}.}
    \label{fig:tigerclaw}
\end{figure}

Our evaluation of the proposed method is conducted within the \textit{Operation TigerClaw} scenario~\cite{dsifirstyear}, which is implemented as a custom map in the \textit{StarCraft II} Learning Environment (\textit{PySC2})~\cite{vinyals2017starcraft}. This platform enables artificial intelligence (AI) agents to engage in the StarCraft II game. \textit{Operation TigerClaw}~\cite{dsifirstyear} presents a combat scenario where Task Force 1–12 CAV is tasked with seizing OBJ Lion by attacking across the Thar Thar Wadi, eliminating the threat force. This objective is depicted in Figure~\ref{fig:tigerclaw}.

In \textit{PySC2}, the scenario is realized by mapping StarCraft II units to their military equivalents, adjusting attributes like weapon range, damage, unit speed, and health. For instance, M1A2 Abrams combat armor units are represented by modified Siege Tanks in tank mode, mechanized infantry by Hellions, among others~\cite{dsifirstyear} (see Appendix \ref{appendix:sc2_tigerclaw_units} for more details).
The Friendly Force consists of 9 Armor, 3 Mechanized infantry, 1 Mortar, 2 Aviation, and 1 Reconnaissance unit. The Threat Force includes 12 Mechanized infantry, 1 Aviation, 2 Artillery, 1 Anti-Armor, and 1 Infantry unit. 

\begin{figure}[htbp]
    \centerline{\includegraphics[width=1.0\columnwidth]{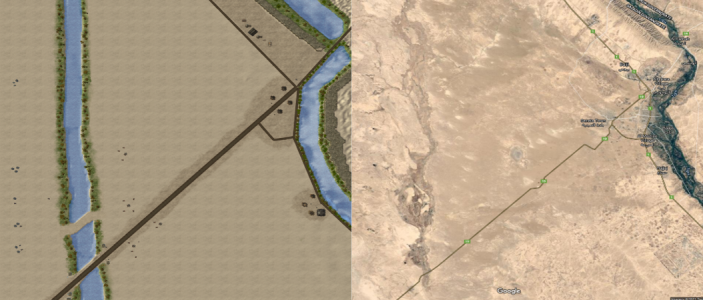}}
    \caption{Satellite view (right) of the area of operations and its representation in StarCraft II (left)~\cite{dsifirstyear}.}
    \label{fig:tigerclaw_sc2_map}
\end{figure}

In terms of terrain, a custom StarCraft II map~\cite{dsifirstyear} was made to depict the TigerClaw area of operations, as shown in Figure~\ref{fig:tigerclaw_sc2_map}.
In this scenario, the Threat Force is controlled by the default in-game StarCraft II AI. They are programmed to automatically move to fixed defensive positions to defend the three main Thar Thar Wadi crossings. Upon detecting the Friendly Force within range, they open fire.
Additionally, the Threat Aviation unit performs a fixed patrol route covering the three crossing points, while also engaging any Friendly Force assets that come into range.

\subsection{COA Generation}

\begin{figure}[htbp]
    \centerline{\includegraphics[width=0.80
    \columnwidth]{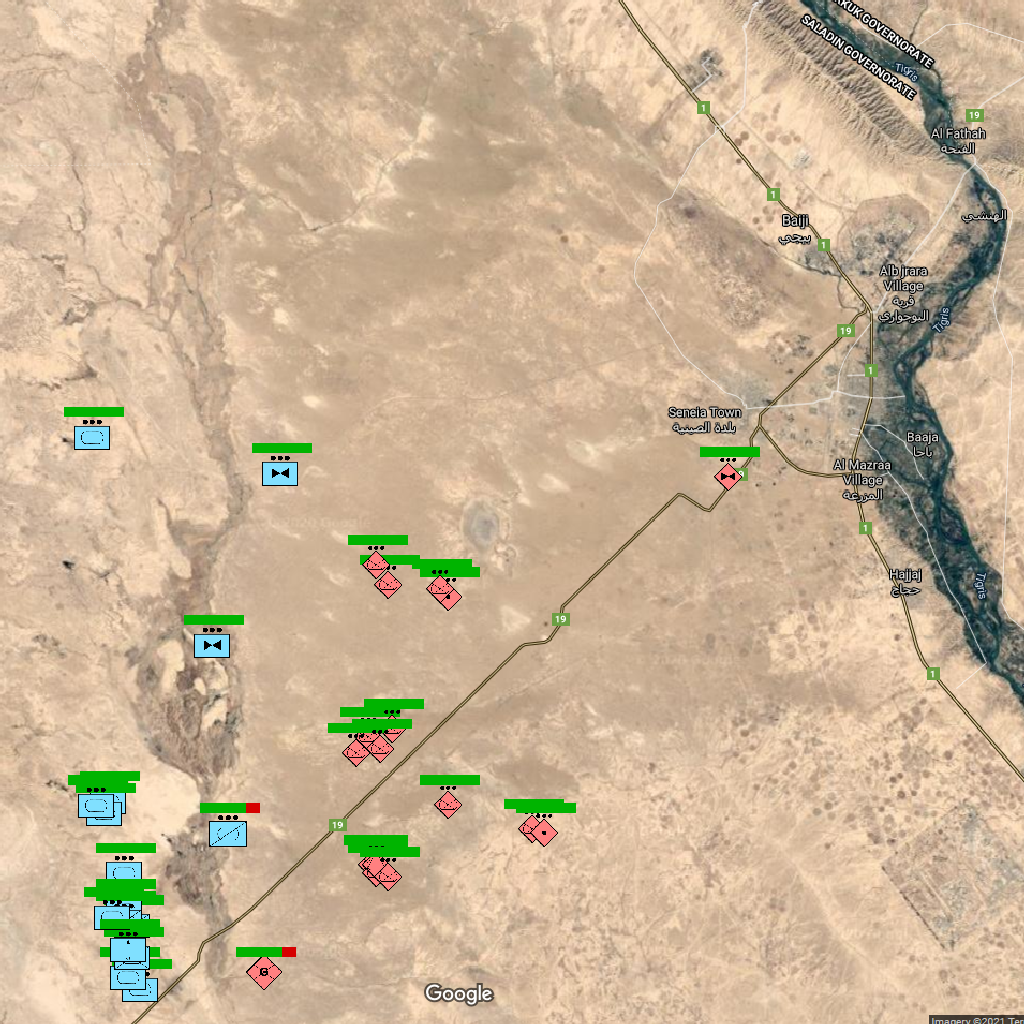}}
    \caption{Example of image given to COA-GPT as part of mission information for experiments with LLMs with vision capabilities. The image overlays force arrangements in a satellite image of the battlefield terrain.}
    \label{fig:frame}
\end{figure}

COA-GPT processes mission objectives and terrain information in text format for all experimental scenarios, which are described as follows:
\begin{itemize}
    \item \textbf{Mission Objective. }``Move friendly forces from the west side of the river to the east via multiple bridges, destroy all hostile forces, and ultimately seize objective OBJ Lion East at the top right corner of the map (coordinates x: 200, y: 89)."
    \item \textbf{Terrain Information. }``The map is split in two major portions (west and east sides) by a river that runs from north to south. There are four bridges that can be used to cross this river. 
    Bridge names and exit coordinates are as follows: 1) Bridge Bobcat (x: 75, y: 26), 2) Bridge Wolf (x: 76, y: 128), 3) Bridge Bear (x:81, y: 179), and 4) Bridge Lion (x: 82, y: 211)."
\end{itemize}
For the experiments using LLM with image processing capabilities, COA-GPT takes as input a frame of a Common Operational Picture (COP) that overlays force arrangements in a satellite image of the terrain, as depicted in Figure~\ref{fig:frame}.

Information regarding Friendly and Threat forces, detailing all assets present in the scenario, is fed to COA-GPT in JSON format:
\begin{itemize}
    \item \textbf{Example Friendly Force Asset: }``\{`unit\_id': 4298113025, `unit\_type': `Armor', `alliance': `Friendly', `position': {`x': 12.0, `y': 203.0}\}"
    \item \textbf{Example Threat Force Asset: }``\{`unit\_id': 4294967297, `unit\_type': `Mechanized infantry', `alliance': `Hostile', `position': {`x': 99.0, `y': 143.0}\}"
\end{itemize}
Mission objective, terrain, and force information are integrated into a single COA generation prompt, as shown in Appendix~\ref{appendix:coa_prompt}, that is then used to query the LLM in the back end.

\begin{figure}[htbp]
    \centerline{\includegraphics[width=1.0\columnwidth]{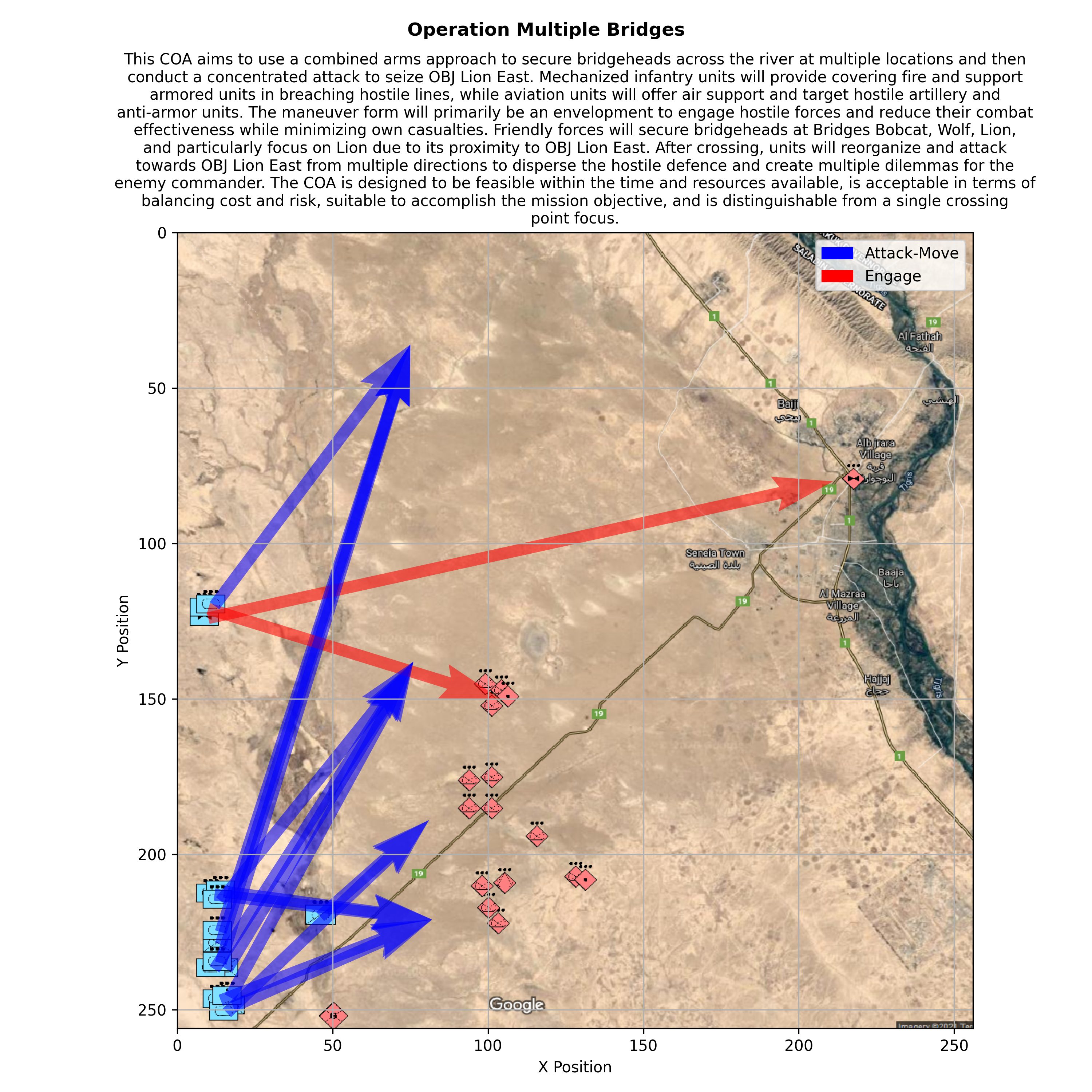}}
    \caption{Sample COA generated by COA-GPT for human review, including a visual representation, mission name, and strategy description.}
    \label{fig:sample_coa}
\end{figure}

For integration with the PySC2 game engine, COA-GPT outputs COAs in JSON format, as detailed in the example COA statement found in Appendix~\ref{appendix:system_prompt}. These task allocations from COA-GPT are translated into specific function calls compatible with the PySC2 engine\footnote{PySC2 Action Definition:~\url{https://github.com/google-deepmind/pysc2/blob/master/pysc2/lib/actions.py}.}, facilitating in-game actions based on the generated COAs.
COA-GPT is programmed with knowledge of the specific game engine functions to command each asset:
\begin{itemize}
    \item \textbf{attack\_move\_unit}(unit\_id, target\_x, target\_y): Directs a friendly unit to move to a specified coordinate, engaging hostile units encountered en route.
    \item \textbf{engage\_target\_unit}(unit\_id, target\_unit\_id, target\_x, target\_y): Orders a friendly unit to engage a specified hostile unit. If the target is out of range, the friendly unit will move to the target's location before engaging.
\end{itemize}

A comprehensive COA encompasses a mission name, a brief strategic description, and designated commands for each asset involved, as defined by the presented game engine functions. During a simulation rollout, each asset strictly follows its initial command issued by COA-GPT at the start of the simulation.
For instance, a command to $engage\_target\_unit(X,Y,coord_x, coord_y)$ would specify unit $X$ to engage target unit $Y$, maintaining this engagement until the battle concludes or a new COA is issued. Similarly, an $attack\_move\_unit(Z,coord_x, coord_y)$ command directs unit $Z$ to move to a specified map coordinate, engaging any hostile units encountered en route.
The PySC2 engine processes these commands at every simulation time step to ensure each asset is correctly controlled.

\subsection{Human Feedback}

As depicted in Figure~\ref{fig:sample_coa}, the COA generated by our system is transformed into a graphical format accompanied by a concise mission summary for presentation to human evaluators. The evaluators can provide textual feedback, which COA-GPT then uses to refine and generate a new COA for subsequent feedback rounds.

To ensure a consistent and fair comparison and evaluation process across all generated COAs, we have standardized the human feedback process specifically for this work. The specific instructions provided in each feedback iteration for the COAs that incorporated human input are as follows:
\begin{itemize}
    \item \textbf{First Iteration: }``Make sure both our aviation units directly engage the enemy aviation unit."
    \item \textbf{Second Iteration: }``Make sure only our Scout unit is commanded to control Bridge Bobcat (x: 75 y: 26) and our other assets (not the aviation) are split in two groups and commanded to move towards both enemy artillery using the \textit{attack\_move} command."
\end{itemize}

After receiving final approval from the human evaluators, COA-GPT proceeds to simulate the scenario multiple times, gathering and compiling various evaluation metrics for analysis.

\subsection{Evaluation Metrics}

The evaluation of the generated COAs is based on three key metrics recorded during the COA analysis: total reward, friendly force casualties, and threat force casualties:
\begin{itemize}
    \item \textbf{Total Reward.} This metric represents the total game score, which is tailored specifically for the TigerClaw mission scenario. It includes positive rewards for strategic advancements and neutralizing enemy units, and negative rewards for retreats and friendly unit losses. Specifically, agents gain +10 points for each unit advancing over bridges and for each enemy unit neutralized. Conversely, they lose -10 points for retreating over a previously crossed bridge and for each friendly unit lost.
    \item \textbf{Friendly Force Casualties.} This metric counts the number of friendly force units lost during the simulated engagement. It reflects the operational cost in terms of friendly unit losses.
    \item \textbf{Threat Force Casualties.} This metric tracks the number of enemy (threat force) units eliminated during the simulation, indicating the effectiveness of the COA in neutralizing the opposition.
\end{itemize}

\subsection{Baselines}

In this study, we benchmark our proposed method against an expert human and the two best-performing published approaches for this scenario: Autocurriculum Reinforcement Learning from a Single Human Demonstration~\cite{waytowich2022learning} and the Asynchronous Advantage Actor-Critic (A3C) algorithm in Reinforcement Learning~\cite{mnih2016asynchronous,waytowich2022learning}:
\begin{itemize}
    \item \textbf{Autocurriculum Reinforcement Learning}~\cite{waytowich2022learning}. Utilizing a single human demonstration, this method develops a tailored curriculum for training a reinforcement learning agent via the A3C algorithm~\cite{mnih2016asynchronous}. The agent, which controls each friendly unit individually, accepts either image inputs from the environment or a vectorial representation detailing unit positions. We evaluate our method against both input modes, referred to as ``Autocurr.-Vec" and ``Autocurr.-Im", respectively, in our results section.
    \item \textbf{Reinforcement Learning.} As detailed in~\cite{waytowich2022learning}, this baseline employs the A3C algorithm, using either images or vector representations as inputs. Unlike the previous method, it does not incorporate a guiding curriculum for the learning agent. Both input modes are evaluated against our method, labeled as ``RL-Vec" and ``RL-Im", respectively, in our results.
    \item \textbf{COA-GPT.} This serves as an ablation study of our proposed approach but without human feedback. Labeled ``COA-GPT" (text-only inputs) and ``COA-GPT-V" (for experiments with vision models using both text and image inputs), this version generates COAs based solely on the initial mission data provided by the C2 personnel, without further human input.
    \item \textbf{COA-GPT with Human Feedback.} This is our fully realized method. Beyond the initial mission information from the C2 personnel, this version also incorporates the impact of human feedback on the COA performance. We assess the changes after the first iteration of feedback, denoted as ``COA-GPT+H1", and after the second iteration, labeled ``COA-GPT+H2" in the results. Similarly, for experiments using the LLM with vision capabilities the experiments are labeled ``COA-GPT-V+H1" and ``COA-GPT-V+H2".
    \item \textbf{Expert Human.} A human expert was tasked to complete the same scenario over 15 trials. Differently from COA-GPT, the human controlled the assets using a mouse a keyboard in real-time throughout the whole game. Similarly, the human user was only able to command attack-move or direct engage commands.
\end{itemize}


\begin{figure*}[htbp]
    \centering
    \begin{subfigure}{0.31\textwidth} 
        \includegraphics[width=\linewidth]{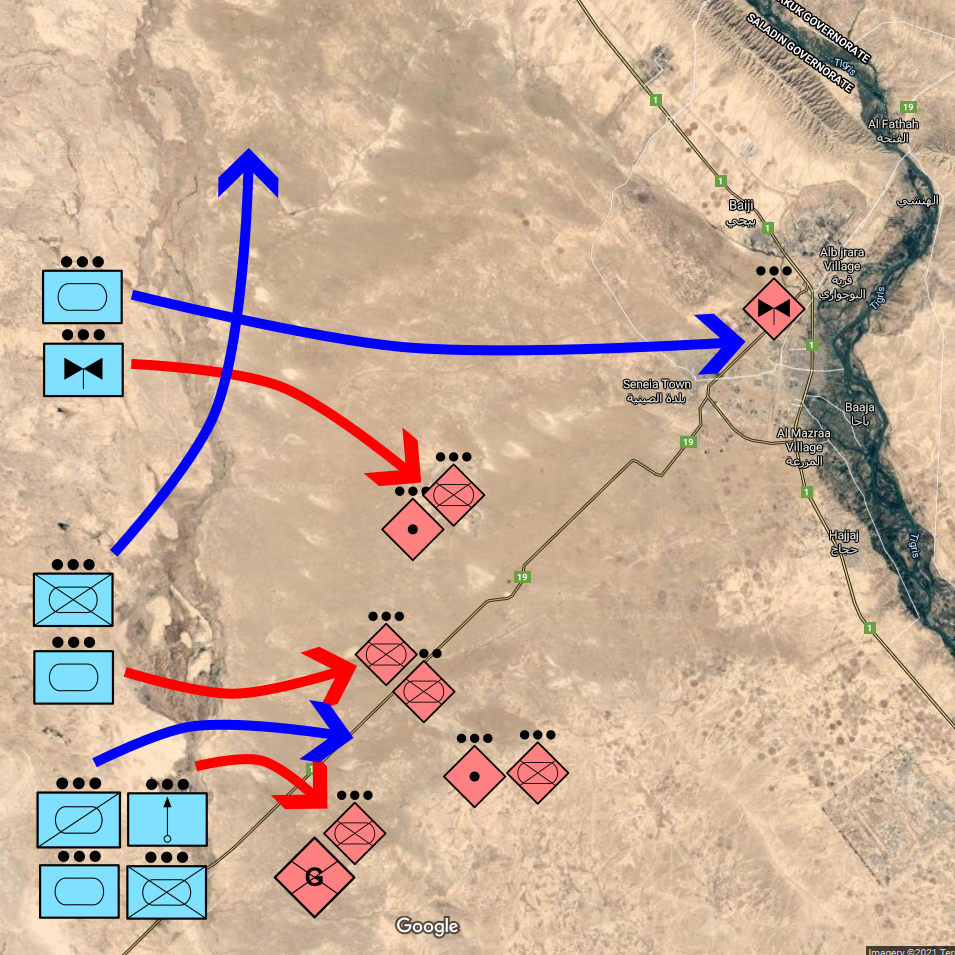}
        \caption{No human feedback.}
    \end{subfigure}%
    \hfill
    \begin{subfigure}{0.31\textwidth}
        \includegraphics[width=\linewidth]{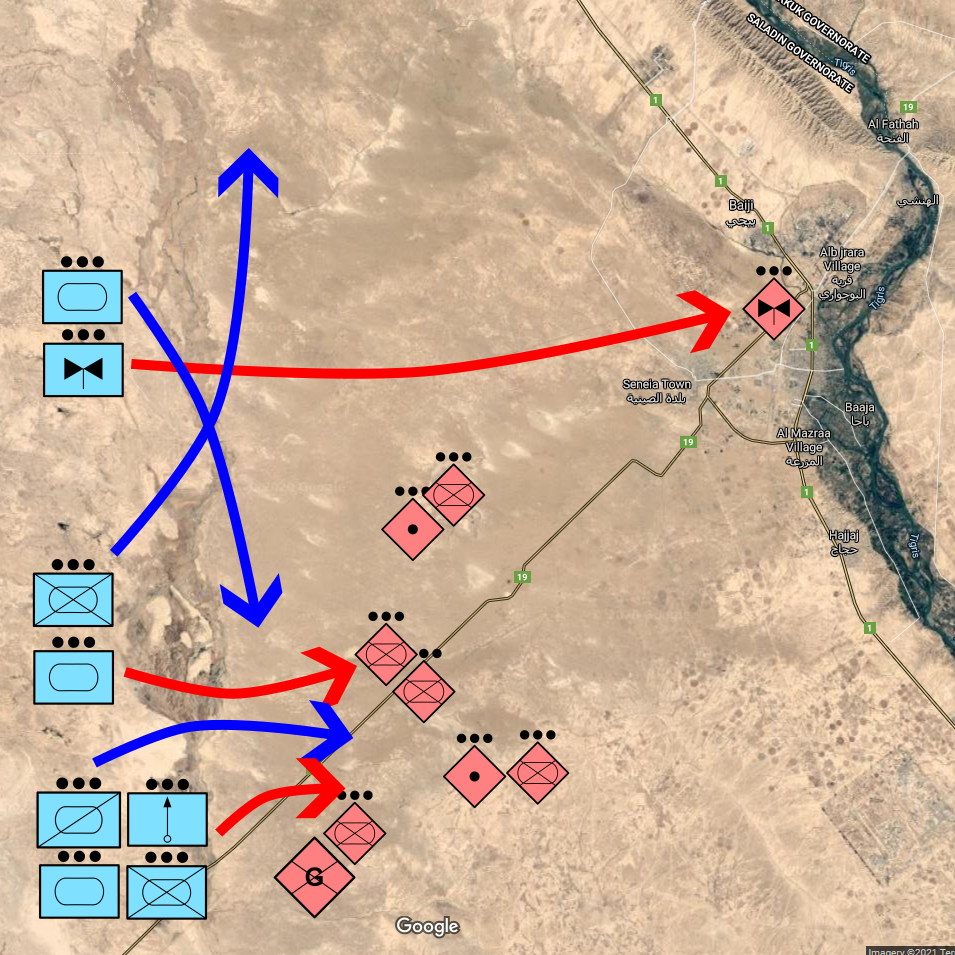}
        \caption{After first human feedback.}
    \end{subfigure}%
    \hfill
    \begin{subfigure}{0.31\textwidth}
        \includegraphics[width=\linewidth]{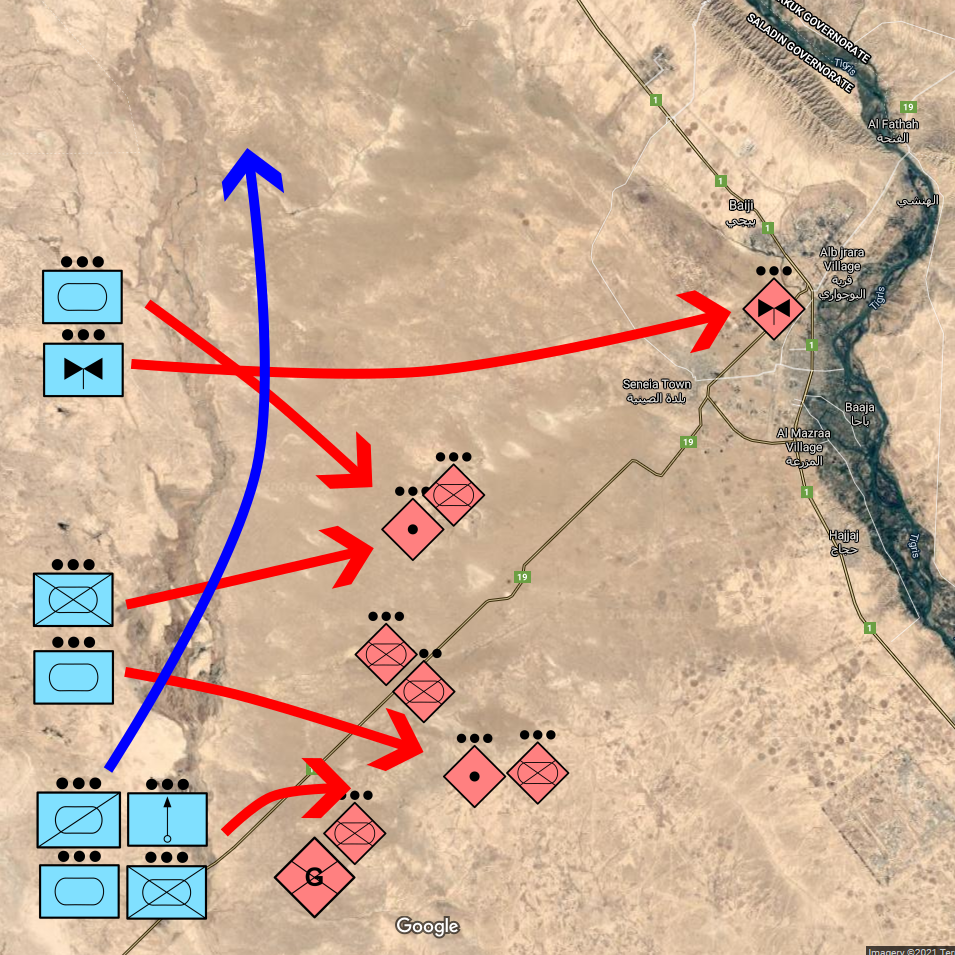}
        \caption{After second human feedback.}
    \end{subfigure}
    \caption{Illustration of the sequential development of COAs influenced by human feedback. (a) The initial COA is depicted, showing movements (blue arrows) of forces across bridges and engagement directives (red arrows) against hostile units, generated without human input. (b) Reflects adjustments made by a human commander specifying direct engagement of hostile aviation by friendly aviation forces. (c) Represents further refinement of the COA where forces are ordered to split to address both enemy artilleries, with instructions for only reconnaissance units to advance to the northern bridge.}
    \label{fig:coa_gpt_example}
\end{figure*}

\section{Experimental Results}

\subsection{Qualitative Results}

To illustrate the full realization of the proposed methodology, Figure~\ref{fig:coa_gpt_example} depicts the evolution of a generated COA in response to human feedback. Initially, the COA generated without human input (Figure~\ref{fig:coa_gpt_example}a) displays a certain configuration for the planned movement of friendly forces across bridges and their engagement with hostile units. After the first round of human feedback (\textit{``Make sure both our aviation units directly engage the enemy aviation unit."}), we see a strategic pivot (Figure~\ref{fig:coa_gpt_example}b); the friendly aviation units are now tasked to engage the enemy's aviation assets directly. This adjustment reflects the human commander’s intent to prioritize air superiority. The second iteration of feedback --- \textit{``Make sure only our Scout unit is commanded to control Bridge Bobcat (x: 75 y: 26) and our other assets (not the aviation) are split into two groups and commanded to move towards both enemy artillery using the \textit{attack\_move} command."} --- as seen in Figure~\ref{fig:coa_gpt_example}c), results in a more nuanced approach: the friendly forces are divided, with specific units tasked to target enemy artillery positions. Additionally, the reconnaissance unit is ordered to secure the northern bridge and the friendly aviation is still tasked to engage the threat aviation, demonstrating that COA-GPT successfully followed the commander's intent conveyed via textual feedback.

\subsection{Quantitative Results}

For a comprehensive evaluation of COA-GPT, we generated five COAs for each method variant and conducted ten simulations for each, totaling 50 evaluation rollouts per baseline.
All results discussed in this section represent the mean and standard deviation calculated across these 50 rollouts for each evaluation metric.
It is important to note that the game of StarCraft II is mostly deterministic~\cite{vinyals2017starcraft} - with stochasticity mostly being injected via weapon speed in the case of our TigerClaw scenario - which implies that the variance present in the results originates from the diversity of the COAs generated by COA-GPT.
The data for the Reinforcement Learning and Autocurriculum Reinforcement Learning baselines are sourced directly from their respective published work~\cite{waytowich2022learning}.
Performance videos of COA-GPT can be seen at the project webpage~\footnote{COA-GPT Project Webpage: \url{https://sites.google.com/view/coa-gpt}.}.

\begin{figure}[htbp]
    \centerline{\includegraphics[width=0.95\columnwidth]{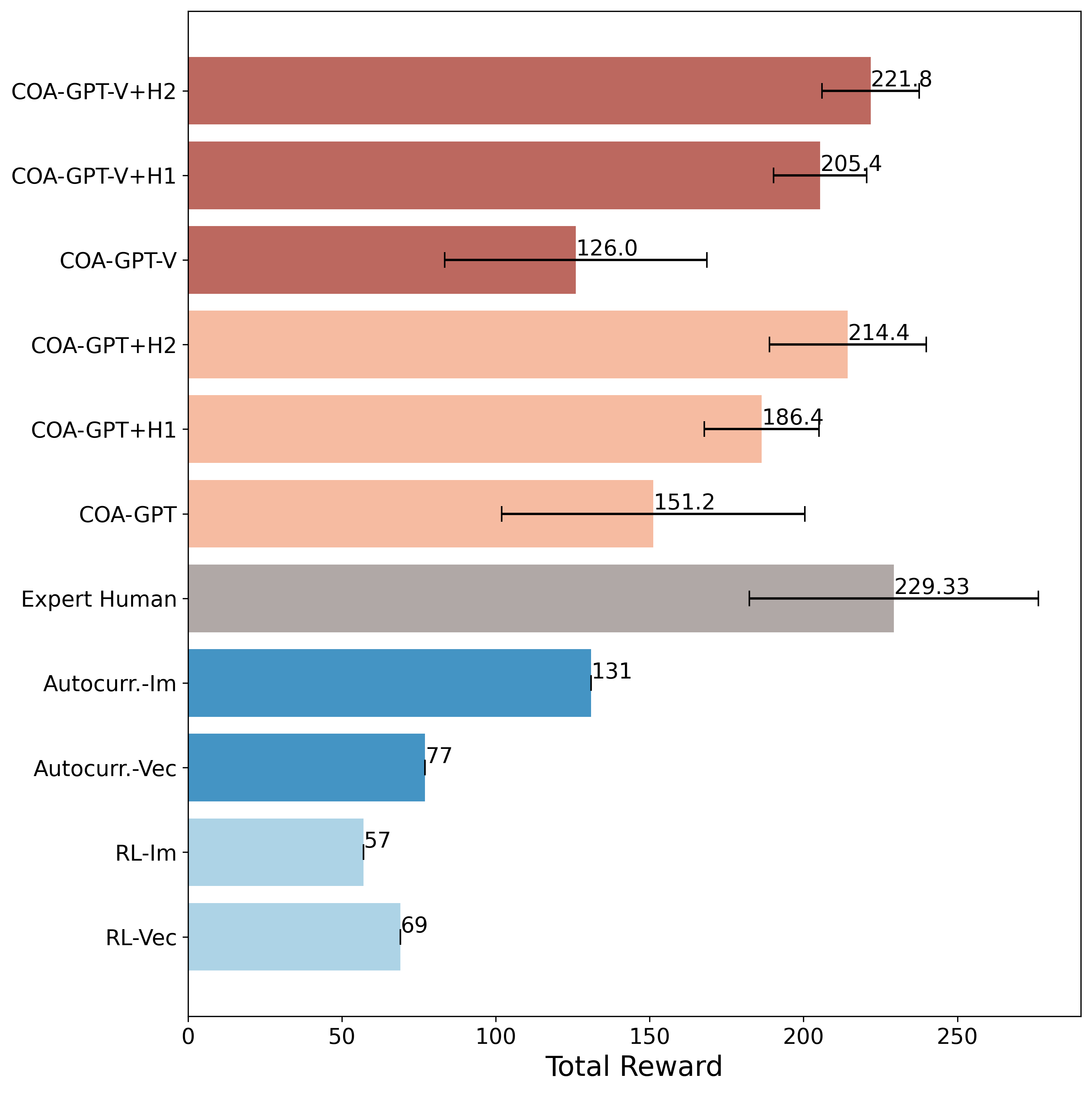}}
    \caption{Comparison in terms of the mean and standard deviation of total reward received across all baselines. COA-GPT generates better COAs after receiving feedback from humans, outperforming all baselines.}
    \label{fig:total_reward_comparison}
\end{figure}

Figure~\ref{fig:total_reward_comparison} presents a comparison of all baseline methods in terms of the average and standard deviation of the total rewards received during evaluation rollouts.
COA-GPT, even in the absence of human interaction and relying solely on textual scenario representations, produces Courses of Action (COAs) that surpass the performance of all AI baselines on average. This includes the previously established state-of-the-art method, Autocurriculum Reinforcement Learning, which utilizes image data.
COA-GPT-V, which also uses a single image as additional input, has equivalent performance compared to the previous baselines. 
Additionally, Figure~\ref{fig:total_reward_comparison} illustrates that the effectiveness of the COAs generated by COA-GPT is enhanced further (as indicated by a higher mean total reward) and exhibits reduced variability (evidenced by a lower standard deviation) when subjected to successive stages of human feedback.
When taking into consideration the performance after human feedback, COA-GPT with vision models (COA-GPT-V+H1 and COA-GPT-V+H2) achieve higher mean total rewards compared to all previous baselines. However, compared to the human expert, the best-proposed model achieves a 3.4\% lower mean score, although it's noteworthy that the human expert's results demonstrate a larger variance.

\begin{figure}[htbp]
    \centerline{\includegraphics[width=0.95\columnwidth]{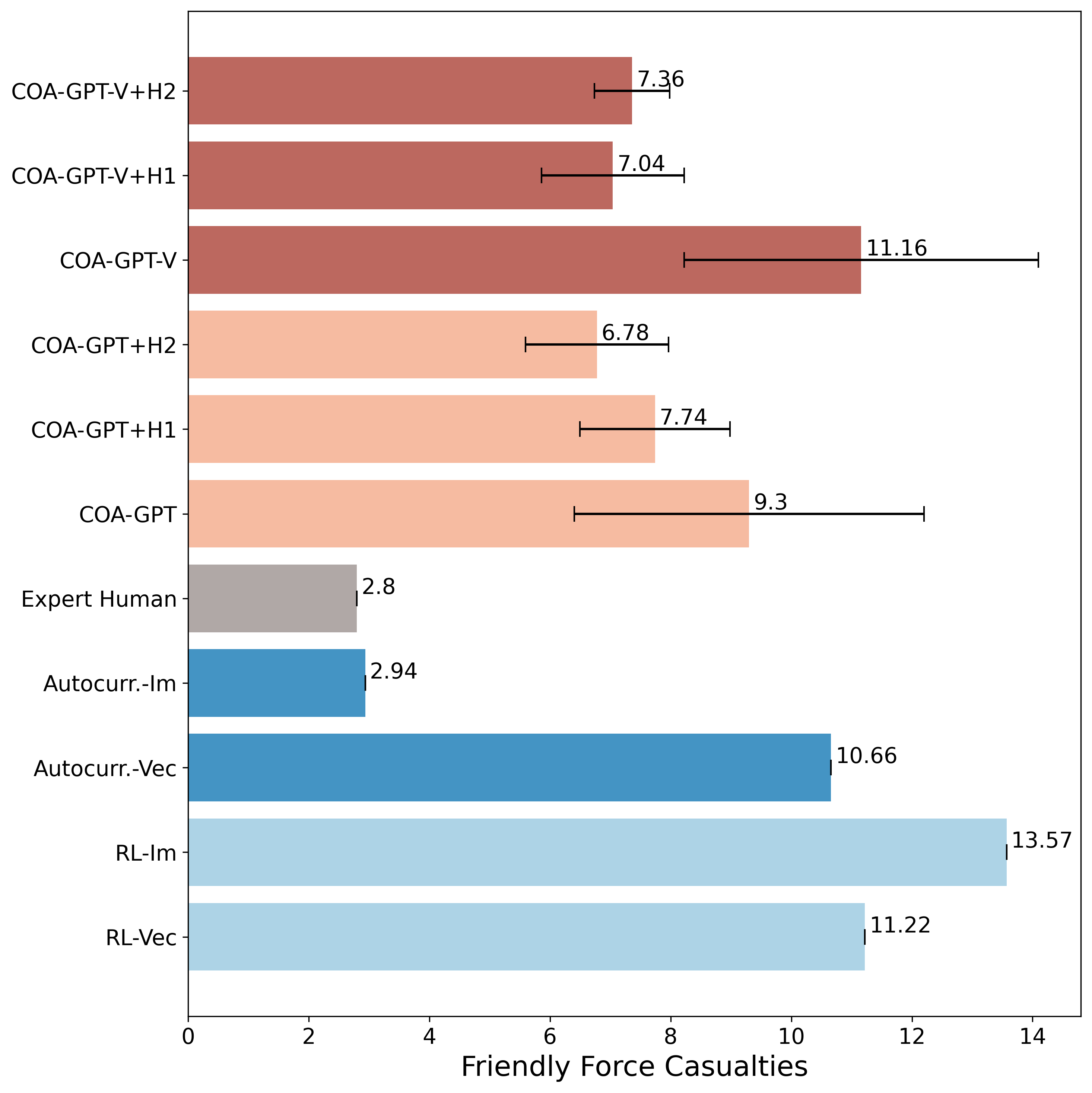}}
    \caption{Comparison in terms of the mean and standard deviation of friendly force casualties across all baselines. Despite human feedback, COA-GPT shows higher mean casualties, suggesting that the precision of control may introduce additional risks.}
    \label{fig:blufor_casualties_comparison}
\end{figure}

\begin{figure}[htbp]
    \centerline{\includegraphics[width=0.95\columnwidth]{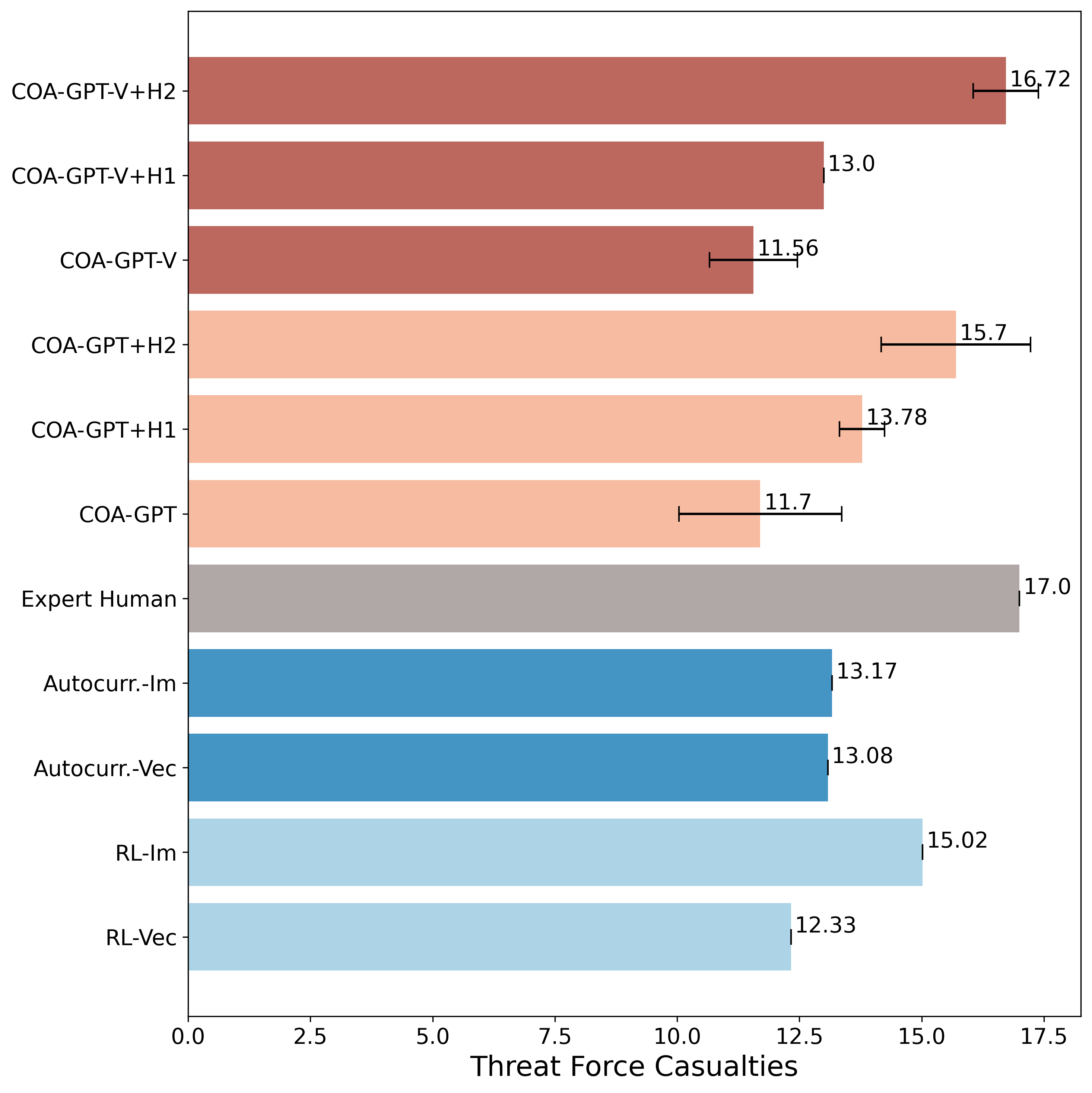}}
    \caption{Comparison in terms of the mean and standard deviation of threat force casualties across all baselines. COA-GPT maintains comparable lethality to other methods, suggesting effective threat engagement irrespective of control granularity.}
    \label{fig:opfor_casualties_comparison}
\end{figure}

Figures~\ref{fig:blufor_casualties_comparison} and~\ref{fig:opfor_casualties_comparison} provide a comparative analysis of the mean and standard deviation of friendly and threat force casualties, respectively, during the evaluation rollouts. In Figure~\ref{fig:blufor_casualties_comparison}, we observe that the COA-GPT and COA-GPT-V, even when enhanced with human feedback (COA-GPT+H1, COA-GPT+H2, COA-GPT-V+H1 and COA-GPT-V+H2 models), exhibits higher friendly force casualties compared to other baselines. This outcome may be linked to COA-GPT’s lower control resolution: offering a single strategic command at the beginning of the episode rather than continuous command inputs throughout the episode, as seen in all the baseline methods. While this approach facilitates a more interpretable COA development process for human operators, it potentially increases casualty rates due to the limited tactical adjustments during COA execution.
The human expert, who was also able to continuously command the assets throughout the episode, achieved the lowest number of friendly casualties and the highest number of hostile casualties, on average.

In contrast, Figure~\ref{fig:opfor_casualties_comparison} will show that the lethality of COA-GPT towards threat forces remains consistent with other baselines, despite operating with lower control resolution. This indicates that COA-GPT and COA-GPT-V variations are capable of matching the effectiveness of other methods in neutralizing threats, even with the potential handicap of less granular command capabilities.

As an important aspect of operational efficiency, we evaluated the time required to generate COAs across all baselines. The Autocurriculum method required an extensive training period, involving 112 thousand simulated battles, equating to 70 million timesteps across 35 parallel workers, to achieve optimal performance with its final trained policy~\cite{waytowich2022learning}.
In contrast, COA-GPT boasts the capability to generate actionable COAs within seconds, demonstrating a significant advantage in terms of rapid deployment.
Moreover, COA-GPT exhibits remarkable flexibility, adapting to new friendly and threat force configurations and changes in terrain without the need for retraining.
This adaptability extends to human commanders who can intuitively adjust COAs generated by COA-GPT.
Such immediate adaptability is not present in the Autocurriculum and traditional Reinforcement Learning methods, which are inherently less flexible due to their reliance on extensive pre-training under fixed conditions before deployment.

\section{Conclusions}

This research presents an advancement in the field of military planning and decision-making through the development of COA-GPT. In the face of increasingly complex and dynamic future battlefields, COA-GPT addresses a critical need in C2 operations for rapid and informed decision-making. Leveraging the power of LLMs, both with text-only and text and images as input, COA-GPT substantially accelerates the development and analysis of COAs, integrating military doctrine excerpts and domain expertise via in-context learning.

The results from the evaluations presented further validate the effectiveness of COA-GPT.
Notably, it demonstrates superior performance in developing COAs aligned to the commander's intent, closely matching expert humans and outperforming other AI methods, including state-of-the-art Autocurriculum Reinforcement Learning algorithms.
While COA-GPT exhibits significant advantages in terms of speed and adaptability, it is enhanced further by the integration of human feedback.
This collaboration leads to COAs that are aligned to human intent, yet are generated much more rapidly, underscoring the tool's potential to transform the pace and efficiency of military planning.

Moreover, the synergy between COA-GPT and human commanders enriches the decision-making process, enabling the exploration of a broader range of operational possibilities and adjustments in real-time. This collaborative approach combines the speed and adaptability of AI with the nuanced understanding and strategic insight of human expertise, facilitating faster, more agile decision-making in modern warfare.

In conclusion, COA-GPT represents a novel approach to enhance military C2 operations.
Its ability to generate actionable COAs within seconds, coupled with its flexible adaptation to new scenarios and intuitive interaction with human commanders, highlights its practical utility and potential for rapid deployment in diverse military scenarios. The development and successful application of COA-GPT pave the way for further innovations in military AI, potentially reshaping how military operations are planned and executed in the future, especially as it demonstrates an improved capacity for dynamic adaptation to changing conditions and enhanced performance through human-AI collaboration.

\section{Future Work}

Future work will extend the application of our methodology beyond the initial scenario presented, exploring its effectiveness across a diverse range of terrain and force configurations.
The proposed method can be adapted to present commanders with multiple COA options, who can further iterate and refine these plans, or be adjusted to fit any established operational workflow.
Additionally, future simulation platforms aim to develop a more complex model of opposition forces. 
This involves transitioning from the current simplistic representation, with fixed patrols and defensive positions, to dynamic simulations that capture the enemy commanders' intentions and strategies, thereby enhancing the realism and strategic depth of our simulations.


\bibliography{refs}
\bibliographystyle{ieeetr}

\appendix

\subsection{COA-GPT System Prompt}\label{appendix:system_prompt}

\begin{lstlisting}
    You are a military commander assistant. Your users are military commanders and your role is to help them develop a military courses of action (COA).

    The military commanders will inform you the mission objective, terrain information, and available friendly and hostile assets before you start developing the COA.
    Given this information, you will develop a number of courses of action (as specified by the commander) so they can iterate on them with you and pick their favorite one.

    For each COA to be complete, every friendly unit needs to be assigned one command from the list below. Hostile units cannot be assigned any command.
        1) attack_move_unit(unit_id, target_x, target_y): commands friendly unit to move to target (x, y) coordinate in the map engaging hostile units in its path.
        2) engage_target_unit(unit_id, target_unit_id, target_x, target_y): commands friendly unit to engage with hostile target unit, which is located at the target (x, y) coordinate in the map. If out of range, friendly unit will move to the target unit location before engaging.
    
    Remember, it is of vital importance that all friendly units are given commands. All generated COAs should be aggregated in a single JSON object following the template below:  
    ```
    {example_coa_statement}
    ```
        
    Here's additional military information that might be useful when generating COAs:
    ```
    {additional_military_info}
    ```

\end{lstlisting}

Variable $example\_coa\_statement$:
\begin{lstlisting}
    {
        "coa_id_0": {
            "overview": <describes overall strategy for this COA, explain why it is feasible (the COA can accomplish the mission within the established time, space, and resource limitations), acceptable (the COA must balance cost and risk with advantage gained), suitable (the COA can accomplish the mission objective), and distinguishable (each COA must differ significantly from the others)."",
            "name": "<name that summarizes this particular COA>",
            "task_allocation": [
                {"unit_id": 4295229441, "unit_type": "Mechanized infantry", "alliance": "Friendly", "position": {"x": 14.0, "y": 219.0}, "command": "move_unit(4295229441, 35.0, 41.0)"},
                {"unit_id": 4299948033, "unit_type": "Aviation", "alliance": "Friendly", "position": {"x": 10.0, "y": 114.0}, "command": "engage_target_unit(4295229441, 3355229433)"},
                <continues for all friendly units, every single one of them: all friendly units need commands>
            ]
        },
        "coa_id_1": {
            <new COA using same template as above>
        }
    }
\end{lstlisting}

Variable $additional\_military\_info$:
\begin{lstlisting}
    - The forms of maneuver are envelopment, flank attack, frontal attack, infiltration, penetration, and turning movement. Commanders use these forms of maneuver to orient on the enemy, not terrain.
- The four primary offensive tasks are movement to contact, attack, exploitation, and pursuit. While it is convenient to talk of them as different tasks, in reality they flow readily from one to another.
- There are three basic defensive tasks - area defense, mobile defense, and retrograde.
\end{lstlisting}

\subsection{COA Generation Prompt}\label{appendix:coa_prompt}

\begin{lstlisting}
    I need to generate a single military course of action to accomplish the following mission objective:
        ```
        {MISSION_OBJECTIVE_TIGERCLAW}
        ```

        The mission is taking place in the following map/terrain:
        ```
        {TERRAIN_TIGERCLAW}
        ```

        The available Friendly and Hostile forces with their respective identification tags, types, and position are defined in the following JSON object:
        ```
        {raw_units_json}
        ```
\end{lstlisting}



Variable $raw\_units\_json$: Data returned from StarCraft II API: $unit\_id$, $unit\_type$, $alliance$, $position$. All details are in the PySC2 package documentation page\footnote{PySC2 Documentation: \url{https://github.com/google-deepmind/pysc2/blob/master/docs/environment.md#observation}.}.

\subsection{StarCraft II to TigerClaw Unit Mapping}\label{appendix:sc2_tigerclaw_units}

Table \ref{tab:unit_mapping} shows a list of StarCraft II units that were used as base to represent the military assets used in the TigerClaw scenario. Units had their speed, range, and damage modified to better approximate their military counterparts~\cite{dsifirstyear}.

\begin{table}[htbp]
\centering
\caption{Original StarCraft II to TigerClaw scenario unit mapping.}
\label{tab:unit_mapping}
\begin{tabular}{@{}l|l@{}}
\toprule
StarCraft II Terran Unit & TigerClaw Unit \\ \midrule
Siege tank (tank mode) & Armor \\
Hellion & Mechanized Infantry \\
Marauder & Mortar \\
Banshee & Aviation \\
Siege Tank (siege mode) & Artillery \\
Reaper & Anti-Armor \\
Marine & Infantry
\end{tabular}
\end{table}

\end{document}